\documentclass[conference]{IEEEtran}
\IEEEoverridecommandlockouts
\usepackage{cite}
\usepackage{amsmath,amssymb,amsfonts}
\usepackage{algorithmic}
\usepackage{graphicx}
\usepackage{textcomp}
\usepackage{xcolor}
\usepackage{amssymb}
\usepackage{subfigure}
 \usepackage{multirow}
\usepackage{balance} 
\usepackage[vlined,boxed,commentsnumbered, ruled, linesnumbered]{algorithm2e}
\usepackage{amsfonts,amssymb}
\usepackage{setspace}

\def\BibTeX{{\rm B\kern-.05em{\sc i\kern-.025em b}\kern-.08em
    T\kern-.1667em\lower.7ex\hbox{E}\kern-.125emX}}
\begin{document}

\title{OIAD: One-for-all Image Anomaly Detection with Disentanglement Learning}
\author{\IEEEauthorblockN{Shuo Wang}
\IEEEauthorblockA{\textit{ Monash University $\&$  CSIRO's Data61  } \\
Melbourne, Australia \\
shuo.wang@csiro.au/monash.edu}
\and
\IEEEauthorblockN{Tianle Chen}
\IEEEauthorblockA{\textit{Monash University} \\
Melbourne, Australia \\
tianlec@student.monash.edu}
\and
\IEEEauthorblockN{Shangyu Chen}
\IEEEauthorblockA{\textit{University of Melbourne} \\
Melbourne, Australia \\
shangyuc@student.unimelb.edu.au}
\and
\and
\IEEEauthorblockN{Surya Nepal}
\IEEEauthorblockA{CSIRO's Data61\\
Sydney, Australia\\
   \ \   surya.nepal@data61.csiro.au\ \ }
\and
\IEEEauthorblockN{Carsten Rudolph}
\IEEEauthorblockA{\textit{Monash University} \\
Melbourne, Australia\\
carsten.rudolph@monash.edu}

\and
\IEEEauthorblockN{Marthie Grobler}
\IEEEauthorblockA{CSIRO's Data61\\
Melbourne, Australia\\
marthie.grobler@data61.csiro.au}
}

\maketitle

\begin{abstract}
Anomaly detection aims to recognize samples with anomalous and unusual patterns with respect to a set of normal data. This is significant for numerous domain applications, such as industrial inspection, medical imaging, and security enforcement. There are two key research challenges associated with existing anomaly detection approaches: (1) many approaches perform well on low-dimensional problems however the performance on high-dimensional instances, such as images, is limited; (2) many approaches often rely on traditional supervised approaches and manual engineering of features, while the topic has not been fully explored yet using modern deep learning approaches, even when the well-label samples are limited. 
In this paper, we propose a One-for-all Image Anomaly Detection system (OIAD) based on disentangled learning using only clean samples. Our key insight is that the impact of small perturbation on the latent representation can be bounded for normal samples while anomaly images are usually outside such bounded intervals, referred to as structure consistency. We implement this idea and evaluate its performance for anomaly detection. Our experiments with three datasets show that OIAD can detect over $90\%$ of anomalies while maintaining a low false alarm rate. It can also detect suspicious samples from samples labeled as clean, coincided with what humans would deem unusual.
\end{abstract}


\section{Introduction}
As a fundamental and challenging machine learning task, anomaly detection aims to recognize images with anomalous and unusual patterns with respect to a set of normal data. Anomaly detection has been applied to a great range of domains, e.g. identification of defective product parts in industrial vision applications \cite{huang2015automated}, fault-prevention in industrial sensing systems \cite{chen2014distributed}, detection of anomalous network activity in intrusion detection systems \cite{ahmed2016survey}, medical image analysis for potential diseases detection \cite{schlegl2017unsupervised,wong2003bayesian}, etc. 
Anomaly detection is achieved by constructing a model of normality and then comparing any input data with that model. Many traditional machine learning techniques have been implemented to detect anomalies in data, such as Bayesian networks, rule-based systems, clustering algorithms, statistical analysis, and Support Vector Machines. 

Anomaly detection techniques can be generally categorized into three types in terms of the availability of data and labels: fully supervised, semi-supervised and unsupervised. In the first scenario, it is assumed that both normal and anomalous data are available for training, and the problem is simplified as a standard classification task. 
In the semi-supervised scenario, only normal data is labeled and available for training, and the goal is to classify new data as either normal or anomalous. The unsupervised scenario or outlier detection is similar to a clustering problem: no labels are given for the training set, which could potentially contain both normal and anomalous data. The goal is to identify the normal cluster while leaving out the outliers. 
There are two key research challenges associated with existing anomaly detection approaches: 
(1) traditional algorithms often perform well on low-dimensional instances but face difficulties when applied to high-dimensional data such as images or speech; 
(2) many approaches often rely on traditional supervised approaches and manual engineering of features, while the topic has not been fully explored yet using modern deep learning approaches. In addition, well-labelled clean and anomalous samples are limited or nonexistent. 

Deep learning omits manual feature engineering and has evolved into a common solution for handling many high-dimensional machine learning tasks. Consequently, this paper aims to investigate the use of deep learning techniques for image anomaly detection. Variational autoencoders (VAE) have achieved state-of-the-art performance in high-dimensional generative modeling. In a VAE, two neural networks – the encoder and the decoder – are pitted against each other. In the process, the decoder learns to reconstruct samples from a low-dimensional latent representation learned by an encoder from a high-dimensional input. 

In this paper, we propose a One-for-all Image Anomaly Detection (OIAD) system based on disentangled learning. Our key insight is that the impact of small perturbations on the latent representation can be bounded for normal samples while anomaly images are usually outside such bounded intervals. This is referred to as called structure consistency. A demonstration is given in Figure 2 and Section III.B. 

Specifically, our algorithm uses only normal instances (also containing suspicious samples) to train a VAE so as for learning disentangled latent representations in an unsupervised manner, where a change in one dimension corresponds to a change in one factor of variation while being relatively invariant to changes in other factors \cite{bengio2013representation}. The structure consistency is then used to determine whether an image is normal or anomalous — that is, whether the reconstruction error based similarity is changed much after small perturbations are added to a given latent representation. If the structure consistency is outside a given threshold interval derived from normal instances, the sample is deemed anomalous. 
We implement this idea of structure consistency and evaluate its performance in anomaly detection. Our experiments with three datasets show that our technique can achieve state-of-the-art performance on standard image benchmark datasets and visual inspection of the most anomalous samples reveals that our method does certainly recognize anomalies.

In the remainder of this paper Sections II explains the background, and Section III describes the system design and our approach. Section IV describes our experimental results. Section V discusses related work, and Section VI concludes the work.

\section{Background}
\subsection{Autoencoders and Variational Autoencoder}
Autoencoders (AEs) are common deep models in unsupervised learning \cite{bengio2013representation}. It aims to represent high-dimensional data through the low-dimensional latent layer, a.k.a. bottleneck vector or code. Architecturally, AEs consist of two parts, the encoder and decoder. The encoder part takes the input $x \in R^d$ and maps it to $z$ (the latent variable of the bottleneck vector). The decoder tries to reconstruct the input data from $z$. The training process of autoencoders is to minimize the reconstruction error. Formally, we can define the encoder and the decoder as transitions $ \tau_1$ and $ \tau_2$:
\begin{equation}
\begin{aligned}
\tau_1(X)\rightarrow Z \\
\tau_2(Z)\rightarrow \hat{X} \\
\tau_1,\tau_2=\underset{\tau_1,\tau_2}{argmin}\left \| X-\hat{X} \right \|^2
\end{aligned}
\end{equation}
The VAEs model has the same structure as the AEs, but is based on an assumption that the latent variables follow some kind of distribution, such as Gaussian or uniform distribution. It uses variational inference for the learning of the latent variables. In VAEs the hypothesis is that the data is generated by a directed graphical model $p(x|z)$ and the encoder is to learn an approximation $q_{\phi} (z|x)$ to the posterior distribution $p_{\theta}(z|x)$. The VAE optimizes the variational lower bound:
\begin{equation}
L(\theta ,\phi ;x) = KL(q_{\phi }(z|x)||p_{\theta }(z)) - \mathbf{E}_{q_{\phi }(z|x)}[log p_{\theta }(x|z)] 
\end{equation}
The left part is the regularization term to match the posterior of $z$ conditional on $x$, i.e., $ q_{\phi }(z|x)$, to a target distribution $ p_{\theta }(z)$ by the KL divergence. The right part denotes the reconstruction loss for a specific sample $x$. In a training batch, the loss can be averaged as:
\begin{equation}
\begin{aligned}
L_{VAE} = \mathbf{E}_{p_{data}(x)}[L(\theta ,\phi ;x) ] \\
=\mathbf{E}_{p_{data}(x)}[KL(q_{\phi }(z|x)||p_{\theta }(z)) ] -\\
\textbf{E}_{p_{data}(x)}[\mathbf{E}_{q_{\phi }(z|x)}[log p_{\theta }(x|z)] ]
\end{aligned}
\end{equation}

\subsection{$\beta$-VAE and Disentanglement Learning}
$\beta$-VAE is a modification of the VAE framework that introduces an adjustable hyperparameter $\beta$ to the original VAE objective: 
\begin{equation}
\mathcal{L} = \mathbb{E}_{q_{\phi }}(log p_{\theta}(x|z))-\beta D_{KL}(q_{\phi }(z|x)|| p_{\theta}(z))
\end{equation}
Well chosen values of $\beta$ (usually $\beta>1$) result in more disentangled latent representations z. 
When $\beta$ = 1, the $\beta$-VAE becomes equivalent to the original VAE framework. It was suggested that the stronger pressure for the posterior $q_{\phi}(z|x)$, to match the factorized unit Gaussian prior $p(z)$ introduced by the $\beta$-VAE objective, puts extra constraints on the implicit capacity of the latent bottleneck $z$ [15]. Higher values of $\beta$ necessary to encourage disentangling often lead to a trade-off between the fidelity of $\beta$-VAE reconstructions and the disentangled nature of its latent code $z$ (see Fig. 6 in [15]). This is due to the loss of information as it passes through the restricted capacity latent bottleneck $z$.

We assume that observations $x^{(i)} \in D; i = 1, cdots, N$ are generated by combining $K$ underlying factors $s = (s_1, cdots, s_K)$. 
These observations are modeled using a real-valued latent/code vector $z \in R^d$, interpreted as the representation of the data. 
The generative model is defined by the standard Gaussian prior $p(z) = N(0; I)$, intentionally chosen to be a factorized distribution, and the decoder $p_{\theta}(x|z)$ parameterized by a neural net. 
The variational posterior for an observation is $ q_{\theta }(z|x)= \prod_{j=1}^{d}N(z_j|u_j(x),\sigma_j^2(x))$, with the mean and variance produced by the encoder, also parameterized by a neural net. 
The variational posterior can be considered as the distribution of the representation corresponding to the data point x. 
The distribution of representations for the entire data set is then given by 
\begin{equation}
q(z)=E_{pdata(x)}[q(z|x)]=\frac{1}{N}\sum_{i=1}^{N}q(z|x^{(i)})
\end{equation}
which is known as the marginal posterior or aggregate posterior, where $pdata$ is the empirical data distribution. A disentangled representation would have each $z_j$ correspond to precisely one underlying factor $s_k$. 
The $\beta-$VAE objective 
\begin{displaymath}
\frac{1}{N}\sum_{1}^{N}[E_{q(z|x^{(i)})}[logp(x^{(i)}|z)]-\beta KL(q(z|x^{(i)})||p(z))]
\end{displaymath}

is a variational lower bound on $E_{q(z|x^(i))}[logp(x^(i))]$ for $\beta \geq 1$. 
Its first term can be interpreted as the negative reconstruction error and the second term as the complexity penalty that acts as a regulariser. 
We may further break down this KL term as $E_{pdata(x)}[KL(q(z|x)||p(z))] = I(x, z)+KL(q(z)||p(z))$; where I(x, z) is the mutual information between x and z under the joint distribution $pdata(x)q(z|x)$. 
Penalizing the $KL(q(z)||p(z))$ term pushes q(z) towards the factorial prior p(z), encouraging independence in the dimensions of z and thus disentangling. 
Penalizing I(x, z), on the other hand, reduces the amount of information about x stored in z, which can lead to poor reconstructions for high values of $\beta$ (Makhzani Frey, 2017). 
Thus making $\beta$ larger than 1, penalizing both terms more, leads to better disentanglement but reduces reconstruction quality. When this reduction is severe, there is insufficient information about the observation in the latent, making it impossible to recover the true factors. 
Therefore, there exists a value of $\beta >1$ that gives the highest disentanglement but results in a higher reconstruction error than a VAE.
\section{Algorithm}
\subsection{Overview}
We propose \textbf{O}ne-for-all \textbf{I}mage \textbf{A}nomaly \textbf{D}etection (OIAD), after a VAE-based disentanglement learning model trained on only normal instances has converged. The encoder has mapped the high-dimensional training data to low-dimensional and disentangled latent representation vector (a.k.a. codes). Given a new sample $x$, small perturbation is added to selected latent codes, so as to generate $m$ morphs by reconstructing $m$ perturbed latent code vectors, called perturbation-based reconstruction morphs (PR-morphs). 
If the autoencoder can well estimate the distribution normal samples and the latent codes are well-selected, then the average reconstruction losses over the final set of morphs will assume low average values for normal samples, and high values otherwise.
\begin{figure}[!htb]
\setlength{\abovecaptionskip}{-0.05cm}
\setlength{\belowcaptionskip}{-0.2cm}
\includegraphics[width=3.5in,height=2.5in]{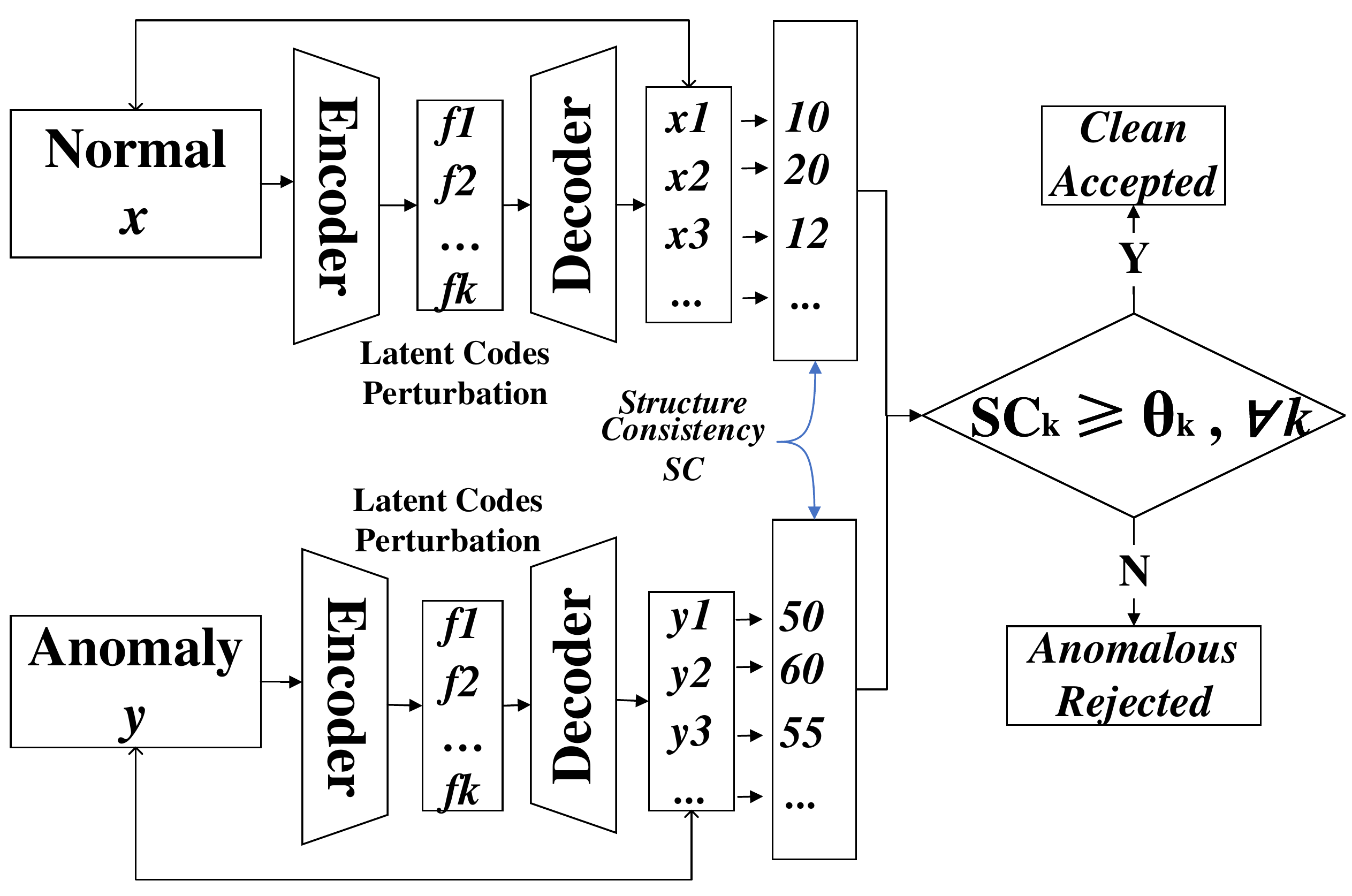}
\centering
\caption{Scheme of OIAD.}
\end{figure}

We expect normal instances are resistant to such perturbation, while anomalies, by contrast, are sensitive to it. Consequently, given a test sample x, if the divergence of the average reconstruction losses evaluation on $m$ PR-morphs, named structure-consistency, is more than a threshold $\alpha$, then it can be inferred that $x$ is anomalous. Our algorithm hinges on this hypothesis, as illustrated in Figure 1.
Two key challenges need to be addressed by OIAD:

1) How to improve the disentanglement performance for efficient PR-morph generation, in an interpretable manner?

2) How to improve the discrimination ability of the detector to recognize anomalies?

To address these issues, two strategies are incorporated in OIAD algorithm:

\subsubsection{Representation Learning with improved disentanglement} We first train a disentangle representation model, Detector-VAE, enhanced from $\beta$-VAE \cite{higgins2017beta}, on a clean dataset (containing some suspicious samples). The objective is to make latent codes more disentangled so that they are easy to be controlled. Such models consist of two components: an encoder $ E: X \rightarrow Z$ and a decoder $ D: Z \rightarrow X’$, where $X$ is the input space and $Z$ is the space of hidden representation. The encoder can map high-dimensional input instance $x$ to disentangled low-dimensional latent codes $z$, i.e., one latent code can only control one certain feature. The decoder is used to reconstruct the input from the low-dimensional latent code $z$. 
Consequently, it is feasible to select and manipulate a given latent code that reveals a specific semantic feature, such as thickness of a digit. The feasibility of feature manipulation rests with the disentanglement. Therefore, strategies are used to improve the disentanglement. Details are described in a later section.

\subsubsection{Detector with fine discrimination ability} Given a sample, we first vary a latent code $i$ for $n$ times to obtain $m$ morphs reconstructed by the decoder. 
We then record the average reconstruction loss evaluation on these $m$ morphs, named structure-consistency. The structure-consistency is used as a resistance indicator $SC^{(i)}$ for code $i$. We can change one or $k$ latent codes simultaneously and obtain a $k$-dimensional resistance vector $\overrightarrow{SC}$ for each instance. We find the resistance ability of normal instances is significantly better than that of anomalous instances, bounded within an obvious interval, as shown in Figure 2. Therefore, a $k$-dimensional threshold configuration $\overrightarrow{\theta_r}$ for all selected $m$ latent codes can be decided on the normal instance to distinguish normal and anomalous instances. An instance that meets $SC^{(i)} > \theta^{(i)},\  \forall\ k$, will be recognized as normal. Otherwise, it will be recognized as anomalous. One obvious example is the digit '1', the structure consistency is strong (i.e. the structure of the image is not changed much when a specific feature, such as thickness, is changed via manipulation of the relevant disentangled latent code), compared with other classes of digits as anomalies.  
\subsection{Demonstration of structure-consistency}
The structure-consistency on the handwritten digits of MNIST \cite{lecun2010mnist} is shown in Figure 2, which is used to demonstrate that the normal images are resistant to such perturbation added to the latent representation (revealed by low reconstruction error). By contrast, anomalies are sensitive to such perturbation.
We consider a detector to distinguish '7' as the normal from other classes of digits as the anomalous. A Detector-VAE is trained on clean '7' images to effectively map instance $x$ to its corresponding latent codes $z$ composed of 20 disentangled latent codes. 
One specific latent code of $z$ is selected to conduct a set of small perturbations, which reveals the degree of crook. The perturbed latent vector can be reconstructed to $x'$ with the help of the decoder. 
Images of the first row are PR-morphs of a clean '7' instance by changing only one specific latent variable ($f_i$) and corresponding structure consistency evaluation. Images of the second row are anomalous images and accordingly PR-morphs and structure consistency evaluation. The structure consistency values of clean instances are bound to obvious intervals, 0.4-0.6 for mean squared error (MSE)-based evaluation, 15-19 for LOSS-based evaluation and 89-93 for SSIM-based evaluation, while structure consistency values of anomalies are out of such intervals.
\begin{figure}[!htb]
    \centering
    \setlength{\abovecaptionskip}{-0.05cm}
    \setlength{\belowcaptionskip}{-0.2cm}
    \includegraphics[width=3.5in,height=2.5in]{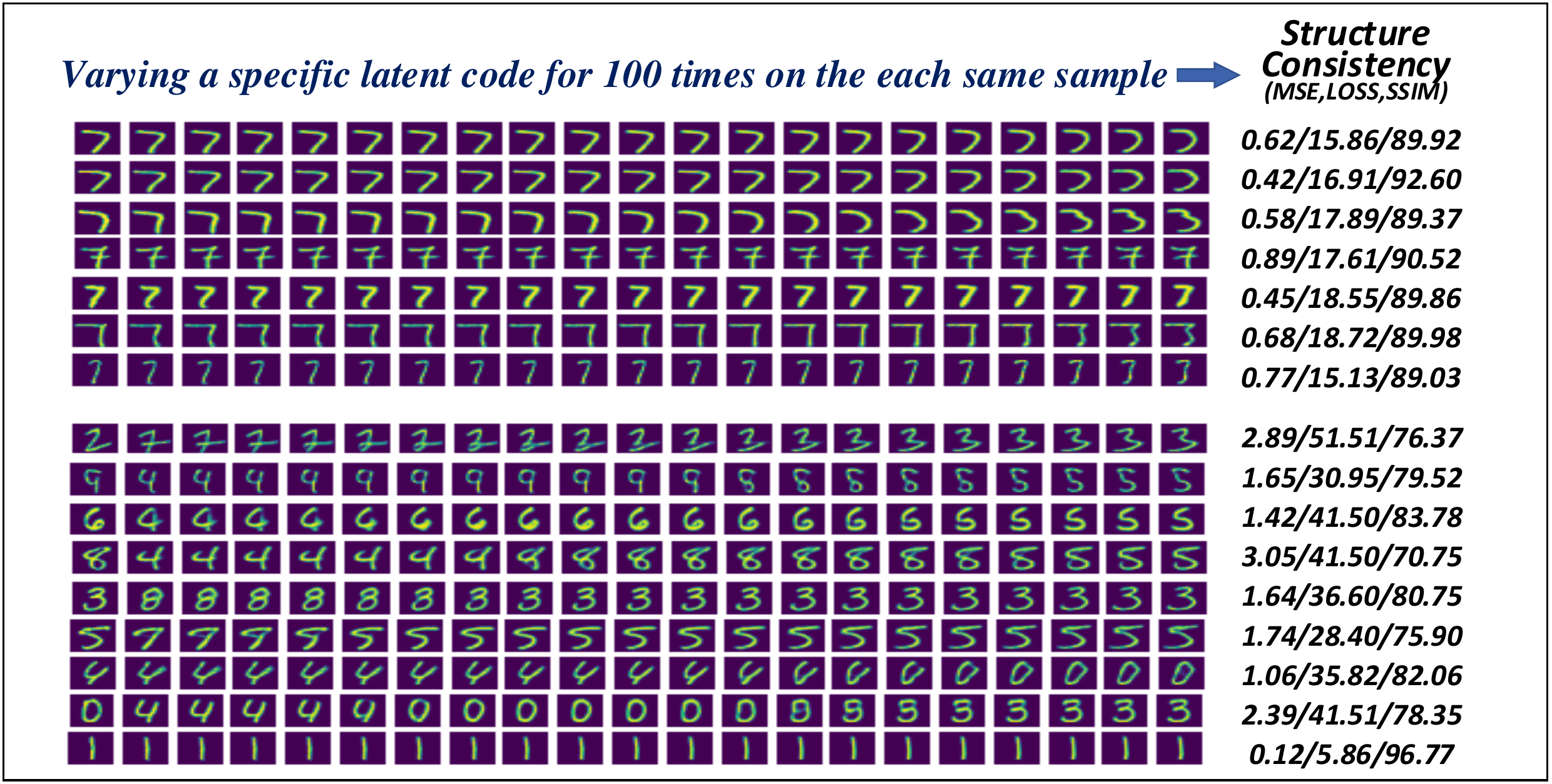}
    \caption{Structure-consistency of MNIST.}
\end{figure}
\vspace{-0.6cm}    
\subsection{Representation with improved disentanglement}
In this section, we aim to achieve a disentangled representation learning task in the unsupervised setting, with no auxiliary information.
VAE-based autoencoders and their variations are commonly applied for disentanglement learning. Specifically, the encoder $E$, parameterized by $q_{\phi }(z|x)$, is trained to convert high-dimensional data $x$ into the latent representation bottleneck vector $z$ in the latent space that follows a specific Gaussian distribution $p(z) \sim N(0, 1)$. The decoder $p_{\theta}(x|z)$ is trained to reconstruct the latent vector $z$ to $x$. The encoder and decoder are trained simultaneously based on the negative reconstruction error and the regularization term, i.e., Kullback-Leibler (KL) divergence between $q_{\phi }(z|x)$ and $p(z)$. The regularization term is used to regularize the distribution $q_{\phi } (z|x)$ to be Gaussian distribution whose mean $\mu$ and diagonal covariance $\sum$ are the output of the encoder. 
In order to obtain good disentanglement in latent codes, we apply two strategies: improving the inner independence of latent codes and adopting complex prior of latent codes. 
Specifically, Total Correlation (TC) \cite{watanabe1960information} is used to encourage independence in the latent vector $z$, as given in Equation 2. 
\begin{equation}
TC(z) = KL(q(z)||\bar q(z)) = Eq(z)[log\frac{q(z)}{\bar q(z)}] 
\end{equation}
As TC is hard to obtain, the approximate tricks used in \cite{kim2018disentangling} is applied to estimate TC. Specifically, a discriminator $D_tc$ is applied to classify between samples from $q(z)$ and $\bar q(z)$. Thus learning to approximate the density ratio is needed for estimating TC \cite{kim2018disentangling}. $ D_{tc} $, parameterized by $\upsilon $, is trained with other components jointly. Thus, the TC term is replaced by the discriminator-based approximation as follows:
\begin{equation}
TC(z) \approx E_{q(z)}[log \frac{D(z)}{1-D(z)}]
\end{equation}

For most existing VAE framework, a standard Gaussian is used as prior for the latent factor, which is suited for modeling of nuisance factors. However, it is demonstrated as both suboptimal and detrimental to performance. Therefore, we additionally apply using long-tail distributions to model relevant factors, as the disentangled latent variables responsible for major sources of variability. Specifically, the VAE is extended to a hierarchical Bayesian model by introducing hyper-priors on the variances of Gaussian latent priors, while maintaining tractable learning and inference of the traditional VAEs \cite{kim2019bayes}. 
For relevant factors, it is necessary to have $p(z_j)$ different from $N(0,1)$. We adopt the complex prior that relaxes the fixed, identical variance assumption for priors $p(z_j)$, defined as follows:
\begin{equation}
p(z|\beta) =\prod_{j=1}^{d}p(z_j|\beta_j) =\prod_{j=1}^{d}N(z_j;0,\beta_j^{-1})
\end{equation}
Here $\beta  > 0$ are the precision parameters to be learned from data. We expect the learned $\beta_j$ to be close to 1 for relevant latent code j. The objective of Detector-VAE is augmented with a TC \cite{watanabe1960information} term to encourage independence in the latent factor distribution and a regularizer $(\beta_j^{-1}-1)^2$ to avoid redundancy in the learned relevant variables, defined as follows:

\begin{equation}
\begin{aligned}
E_{ q_{\phi } (z|x^{(i)})}[log p_{\theta} (x^{(i)}|z)+\\ \sum_{1}^{d}E_{ pd(x)}[ KL( q_{\phi } (z|x^{(i)})||N(z_j;0,\beta_j^{-1}))] \\
+  \gamma L_{TC}+ \eta (\beta_j^{-1}-1)^2
\end{aligned}
\end{equation}
Note that this is also a lower bound on the marginal log likelihood $E_{p(x)}[log p(x)]$. 
The first part reveals the reconstruction error, denoted by $L_R$, evaluating whether the latent bottleneck vector z is informative enough to recover the original instance. $L_R$ can be defined as the $l_2$ loss between the original instance and the reconstructed instance. 
The second part is a regularization term, denoted by $L_{KL}$, to push $ q_{\phi } (z|x)$ to match the prior distribution $p(z)$.
The third part is the TC term, denoted by $L_{TC}$, to measure the dependence for multiple random variables. 
The last one is the regularizer. The trade-off parameter $\eta $ is a proxy to control the cardinality of relevant factors; small $\eta $ encourages more relevant factors.

The parameter $\phi $ of encoder $ q_{\phi } (z|x)$ is then trained by $L_{KL}$, $L_R$ and $L_{TC}$ in terms of $-\nabla_{ \phi }( L_{KL} +L_{R}+ \gamma L_{TC}) $. 
The parameter $\theta $ of decoder is updated in terms of $-\nabla_{\theta }(L_R) $. 
The parameter $ \upsilon $ of TC-discriminator is updated in terms of $-\nabla_{\upsilon }(L_T) $, i.e. $-\nabla_{\upsilon }\frac{1}{2|B|}[\sum_{i \in B}log (D_\upsilon (z^{(i)})+\sum_{i \in B'}log (1-D_\upsilon (permutedim(z'^{(i)}))] $. Here, the permutedim function is to random permutate on a sample in the batch for each dimension of its $z$, similar with \cite{kim2018disentangling}.

\subsection{Detector with fine discrimination ability}
The indicator for anomaly detection should easily differentiate normal and anomalous instances, be feasible and stable to conduct. The discrimination ability of the detector depends on the naturality of the morphs and the accuracy of reconstruction error evaluation. Consequently, we apply two strategies: natural morph generation and feature-wise reconstruction evaluation. 

\subsubsection{Natural morph generation} 
 The initial step is to find the normal value range of each code on the normal validation set, then the morphs are produced via manipulating each code within its normal value range. As the latent codes are disentangled, independent (all from $N(0,1)$) and have semantic meaning, some latent codes that are well distinguished will be selected and their normal ranges can be empirically decided in a human interpretable manner on a validation set. 
 To obtain the morphs by feature manipulation, we can incrementally add/reduce a fixed value on the original learned latent codes within the normal range. However, the modified latent vector maybe not be on the manifold of normal instances. If that happens, an unnatural instance will be reconstructed by the decoder. Hence, we conduct an iterative stochastic search to make the morphs on the manifold by adding natural noise. Specifically, we increase the search range by $\Delta r$ within which the perturbation for a certain latent code $\Delta z_i$ is randomly sampled ($B$ samples for each iteration) until we produce $N$ natural latent code with the value in the normal value range to reconstruct $N$ natural morphs. We then evaluate the structure-consistency for this latent code based on the average reconstruction error. 
Given a targeted classifier, we decide a structure-consistency threshold for each latent code on a validation set containing only normal instances. 
 The threshold of resistance is decided for each factor using the $\alpha$-fractile and $1-\alpha$-fractile points on the validation set. 
 
\subsubsection{Feature-wise structured reconstruction evaluation}
It is always hard to decide a well-grained threshold using the pixel-wise reconstruction error, as too fine-grained threshold leads to high false-negative error (normal samples to be detected as anomalies) and too coarse-grained threshold causes the high true-negative error (anomalies to be detected as the normal).
Instead of the pixel-wise squared error, we use a higher-level representation of the images, a feature-wise structured similarity (SSIM), to measure the similarity. 
Inspired by \cite{gatys2015neural,larsen2015autoencoding}, the feature-wise metric derived from the properties of images learned by the discriminator, a.k.a. style error or content error, is used as a more abstract reconstruction error to better measure the similarity between the original instance and reconstructed ones, aiming to improve the utility of reconstructed instance. That is, we can use learned feature representations in the discriminator of a pre-trained GAN as the basis for the VAE reconstruction objective or we can train a VAE-GAN structure \cite{larsen2015autoencoding} synchronously.
In this work, we apply three different structure consistency evaluation metrics: mean squared error (MSE)-based, Loss-based and SSIM-based to measure the difference between original sample and its reconstructed sample via Detector-VAE, respectively. The loss here combines the reconstruction error and the KL divergence values together. The smaller MSE and Loss-based evaluations are, the better the reconstruction will be, SSIM-based evaluation in contrast.  

\section{Experiments}
In this section, we demonstrate the efficacy of OIAD for image anomaly detection on three image datasets, compared to competing methods. We show experimental evidence that OIAD outperforms non-parametric as well as available deep learning approaches on controlled experiments where ground truth information is available. Additionally, OIAD may be implemented on large, unlabeled data to detect anomalous samples that coincide with what humans would deem unusual.
\subsection{Datasets}
The performance of OIAD is evaluated on three popular image datasets. 
(1) MNIST \cite{lecun2010mnist} consists of $28\times28$ grayscale handwritten digit images from 10 classes, i.e., digit 0-9 and has a training set of 55000 instances and a test set of 10000 instances. 
(2) Fashion MNIST (FMNIST) dataset \cite{xiao2017fashion}, consists of a training set of 60000 examples and a test set of 10000 examples. Each example is a $28\times 28$ grayscale image, associated with a label from 10 classes.
(3) CIFAR-10 \cite{Krizhevsky09learningmultiple} consists of 60000 color images of size $32\times 32$, divided into 10 classes with 6000 images per class. There are 50000 training images and 10000 test images.



\subsection{ Methods, competitor and setups }
For MNIST, FMNIST and CIFAR-10, $70\%$ normal examples from a given class are chosen for training class-unique Detector-VAE. 
In the ground truth information available case, we randomly select  $20\%$ normal images (named CLE, labeled 0) and the corresponding same number of anomalous samples (named ANO, labeled 1, considering samples from other classes as anomalies), respectively. These datasets are used to test the efficiency of the OIAD. An additional $10\%$ normal instances are chosen as the validation data (named VAL) to decide thresholds. 
In the unlabeled case, we randomly select $50\%$ of a given class image (named TEST) to train Detector and use the rest to detect anomalous samples and then to be confirmed by humans.
We normalized the data between 0 and 1 instead of [0, 255] for simplicity. 
Table I shows the architectures of the Detector-VAE for MNIST, FMNIST, and CIFAR-10.

We select 20-dimensional latent codes to manipulate MNIST, FMNIST, and 40-dimensional for CIFAR-10. We generate $100$ morphs for each latent factor of instance by varying each selected latent code for 100 times within according normal value range. The morphs were then evaluated the average reconstruction error compared with their original samples. 
For MNIST, FMNIST and CIFAR-10, we decided the resistance threshold vector, respectively, using the $40\%$-fractile and the $1-40\%$-fractile point on the validation set VAL.
This means each detector mistakenly rejects no more than $40 \%$ normal instances in the validation set, i.e., $\alpha=40\%$. Other default hyper-parameters are given as follows: $\gamma=40,\ \lambda_{lkd}=0.1,\ \eta=0.1$.

\begin{table}[!htb]
\setlength{\abovecaptionskip}{-0.05cm}
\setlength{\belowcaptionskip}{-0.2cm}
\caption{The network structures}
\scriptsize
\centering
\label{tab:my-table}
\begin{tabular}{|c|c|l|}
\hline
\textbf{\textbf{Encoder }} & \textbf{\textbf{Decoder}} & \textbf{TC-Discriminator} \\ \hline
Conv.ReLU 4*4*32 stride 2 & Dense.ReLU 128/512        & 6*Linear.ReLU 1000        \\ 
Conv.ReLU 4*4*32 stride 2 & Dense.ReLU 4*4*64         & Linear.ReLU 2             \\ 
Conv.ReLU 4*4*64 stride 2 & Conv.ReLU 4*4*64 stride 2 &                           \\ 
Conv.ReLU 4*4*64 stride 2 & Conv.ReLU 4*4*32 stride 2 &                           \\ 
Dense 128     & Conv.ReLU 4*4*32 stride 2 &                           \\ 
              & Conv.ReLU 4*4*1stride 2   &                           \\ \hline
\end{tabular}
\end{table}

We tested the performance of OIAD against three commonly used non-parametric anomaly detection approaches: 
1) KDE with a Gaussian kernel \cite{parzen1962estimation}. 
2) One-class support vector machine (OC-SVM) \cite{scholkopf2001estimating}  with a Gaussian kernel ($\nu = 0.1$). 
3) Gaussian mixture model (GMM). We allowed the number of components to vary over $\{2,3,\cdots,20\}$ and selected suitable hyper-parameters by evaluating the Bayesian information criterion. 
Note that feature dimensionality is reduced before conducting anomaly detection via PCA \cite{pearson1901liii}, varying the dimensionality over $\{20, 40, \cdots, 100\}$. At last, the best performance on a small holdout set is used for evaluation. The experimental setup of competing methods generally follows OC-SVMs \cite{scholkopf2001estimating}. 
We also report the performance of deep anomaly detection approaches: ordinary VAEs and DCAEs based detectors. For the VAE and DCAE, we scored according to reconstruction losses, interpreting a high loss as indicative of a new sample differing from samples seen during training. 
In both DCAEs and VAEs, we use a convolutional architecture similar to that of DCGAN \cite{radford2015unsupervised}. 
\subsection{Evaluation on labeled image}
We first evaluate the performance of the anomaly detection for varying structure-consistency thresholds, i.e. $\alpha$. The Detector-VAE is trained on data from a single class $y_c$ from MNIST. Then we evaluate the performance of OIAD on 5000 items randomly selected from the test set, which contains samples from all classes, considering $y \neq y_c$ as anomalous. We expect a high average reconstruction error allocated to images from anomalous classes and a low score to the normal class. 
Here, we test on both CLE and ANO datasets, respectively. The correct decision is that anomalies are recognized as 1 while 0 for normal ones. The anomaly detection accuracy (legend as TP) is the proportion of anomaly instances in ANO to be recognized as an anomaly, i.e. True-Positive. The normal detection accuracy (legend as TN) is the proportion of normal instances in CLE to be recognized as normal, i.e. True-Negative. The overall detection accuracy is the proportion of all correctly detected instances in both ANO and CLE. 
The overall results are shown in Figure 3.
The first row of Figure 3 is the result of training an anomaly detector for digit 1 of MNIST. When varying the threshold, the TP increases while TN decreases. 
We observe that even for a small resistance threshold, e.g. $\alpha=7\%$ for all MSE-based, LOSS-based and SSIM-based structure consistency evaluation, it can detect $100\%$ anomalies. SSIM-based structure consistency outperforms other metrics, which is also illustrated by the first subfigure in Figure 4. 
\begin{figure}[!htb]
\setlength{\abovecaptionskip}{-0.05cm}
\setlength{\belowcaptionskip}{-0.2cm}
\includegraphics[width=3.5in,height=4.8in]{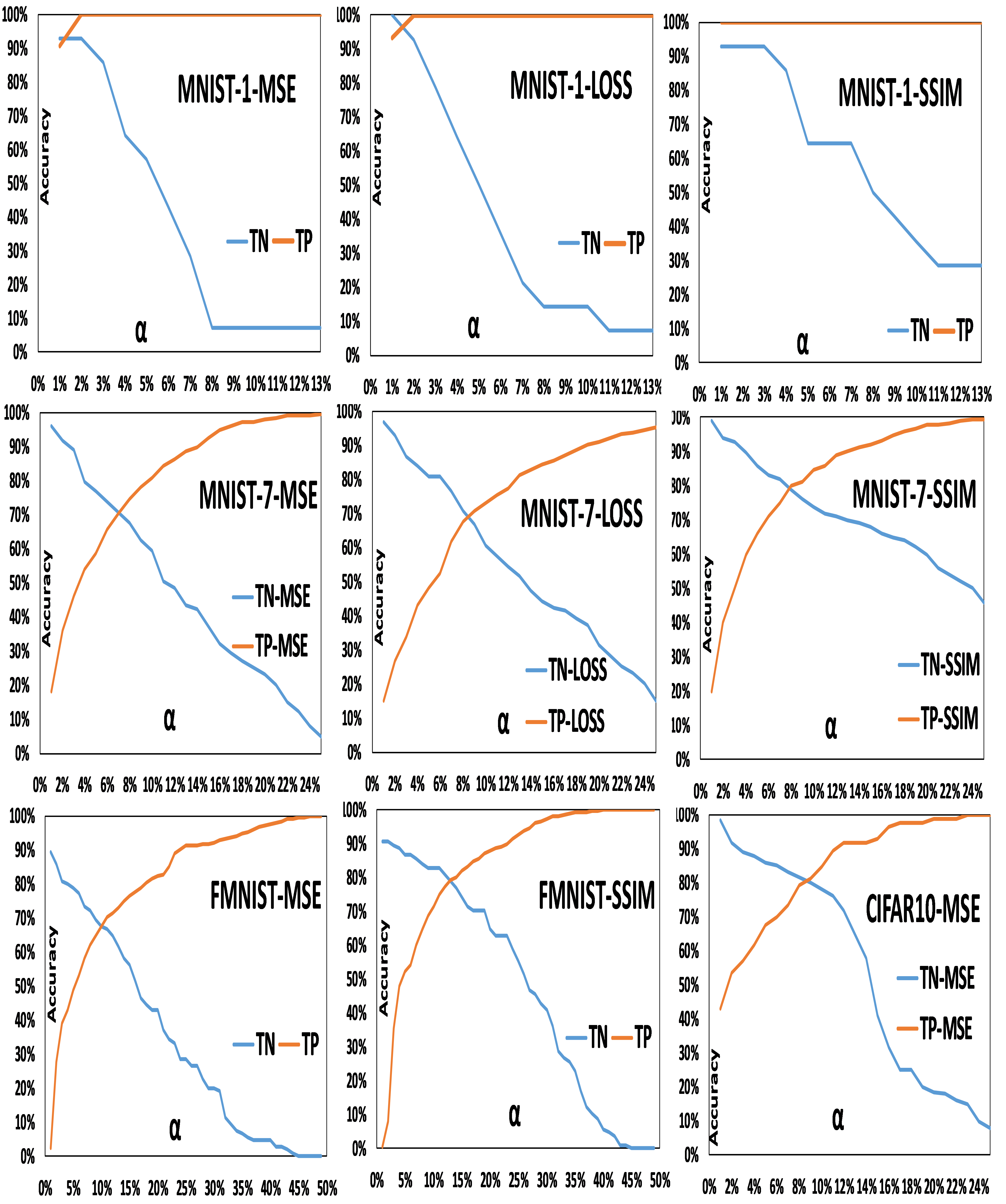}
\centering
\caption{Detection performance with varying resistance thresholds on labeled images.}
\end{figure}
\begin{figure}[!htb]
\setlength{\abovecaptionskip}{-0.05cm}
\setlength{\belowcaptionskip}{-0.2cm}
\includegraphics[width=3.5in,height=2.3in]{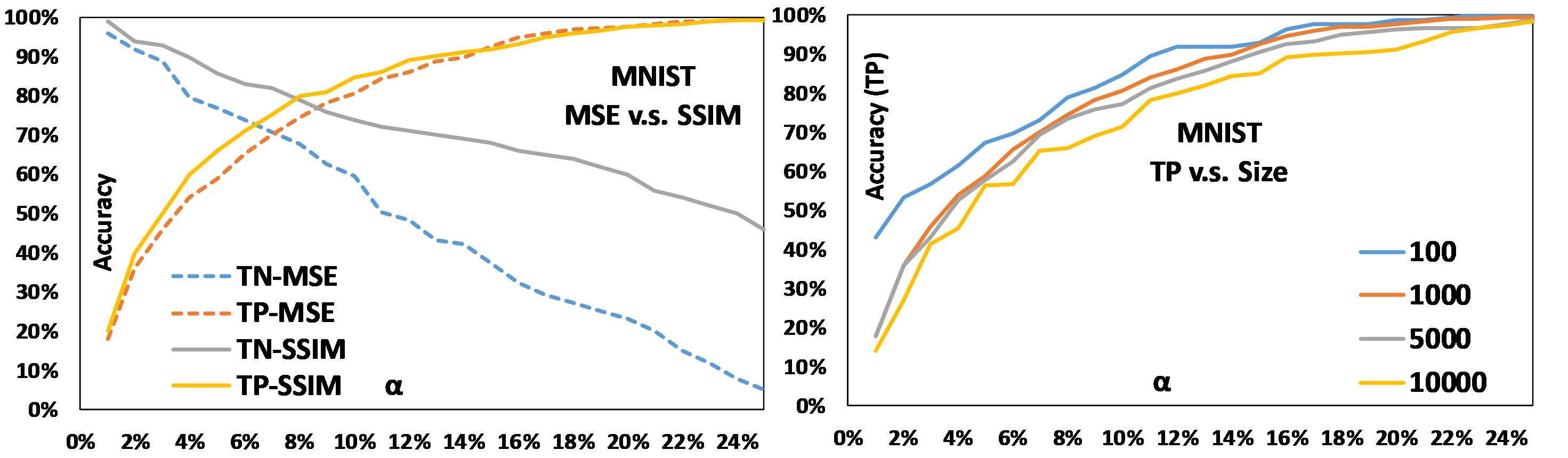}
\centering
\caption{Detection performance with varying reconstructure error metrics and size of validation datasets.}
\end{figure}

The second row of Figure 3 is the result of training an anomaly detector for digit 7 of MNIST. When varying the threshold, the TP increases while TN decreases as well. Due to the diversity of the digit 7 writing styles, the TP is not as high as the first detector. A high TP rate is achieved at the expense of reducing TN. However, we observe that it is feasible to find a small resistance threshold, e.g. $\alpha=13\%$ for SSIM-based structure consistency evaluation, it can detect more than $90\%$ anomalies while maintaining $70\%$ TN rate. Overall, the performance of SSIM-based structure consistency outperforms other metrics. 
The last row of Figure 3 demonstrates the performance of OIAD on FMNIST and CIFAR-10. 
On FMNIST, we train a trousers-anomaly detector, considering other types of icons as anomalies. On CIFAR-10, we train a dog-anomaly detector and cat-anomaly detector respectively, considering other types of images as anomalies. These aforementioned findings are confirmed again. It is interesting to find that the distinguishability of the OIAD is better on the more complex image datasets. For instance, a small resistance threshold, e.g. $\alpha=11\%$ for SSIM-based structure consistency evaluation, can detect more than $90\%$ anomalies while maintaining $72\%$ TN rate. 
Additionally, we also demonstrate the effect of the size of the dataset to decide the threshold. As shown in the second subfigure in Figure 4, the relatively small size of the dataset can provide good thresholds that can achieve a high TP rate. 

In Table II, we report the AUC-ROC on each class-unique anomaly detection scenarios. In these controlled experiments we highlight the ability of OIAD to outperform traditional methods at the task of detecting anomalies in a collection of high-dimensional image samples. Overall, OIAD shows the best performance compared to all comparisons. Note that we achieved such high accuracy without any anomalies required and only based on threshold vectors that are easy to be decided experimentally. 

\begin{table}[!htb]
\setlength{\abovecaptionskip}{-0.03cm}
\setlength{\belowcaptionskip}{-0.1cm}
\scriptsize
\caption{AUC-ROC of anomaly detection on MNIST/CIFAR-10}
\label{tab:my-table}
\begin{tabular}{|l|l|l|l|l|l|l|l|}
\hline
                       & $y_c$   & KDE   & OC-SVM & GMM   & VAE   & DCAE  & OIAD  \\ \hline
\multirow{3}{*}{MNIST} & 1       & 0.999 & 1      & 0.999 & 0.998 & 0.992 & 1     \\ \cline{2-8} 
                       & 7       & 0.934 & 0.952  & 0.937 & 0.896 & 0.941 & 0.966 \\ \cline{2-8} 
                       & Average & 0.966 & 0.975  & 0.968 & 0.947 & 0.966 & 0.983 \\ \hline
\multirow{3}{*}{CIFAR} & cat     & 0.521 & 0.523  & 0.446 & 0.666 & 0.546 & 0.814 \\ \cline{2-8} 
                       & dog     & 0.44  & 0.516  & 0.504 & 0.494 & 0.642 & 0.849 \\ \cline{2-8} 
                       & Average & 0.481 & 0.512  & 0.475 & 0.58  & 0.594 & 0.832 \\ \hline
\end{tabular}
\end{table}
\subsection{Evaluation on unlabeled image}
Since a high TP rate is achieved at the expense of reducing the TN, in this section, we will investigate these samples labeled as normal but to be detected as anomalies. 
We also demonstrate the performance of OIAD in a practical setting where no ground truth information is available. 
For this, we first trained a Detector-VAE on a class-specific dataset that is clean but possibly contains anomalies. We then used the OIAD to find the most anomalous images within the corresponding validation sets containing 1000 images. The thresholds are decided by using the $\alpha$-fractile and $1-\alpha$-fractile points. 
We consider three scenarios: digit 7 anomaly detector on ‘7’ instances only of MNIST, trousers anomaly detector on trousers only of FMNIST and dog/cat anomaly detector on dog/cat only of CIFAR-10. 
The images that are outside the decided intervals on three scenarios are shown in Figures 5, 6 and 7, respectively. 
Among the images recognized as anomalous in these three scenarios, approximately $100\%$ of them are deemed to be unusual by humans. It shows that our method has the ability to discern the normal from the unusual samples. We infer that OIAD is able to incorporate many significant properties of an image. Samples that are assigned good structure consistency scores are in line with a classes’ ‘Ideal-Form’.

\begin{figure}[!htb]
\setlength{\abovecaptionskip}{-0.05cm}
\setlength{\belowcaptionskip}{-0.2cm}
\includegraphics[width=2.2in]{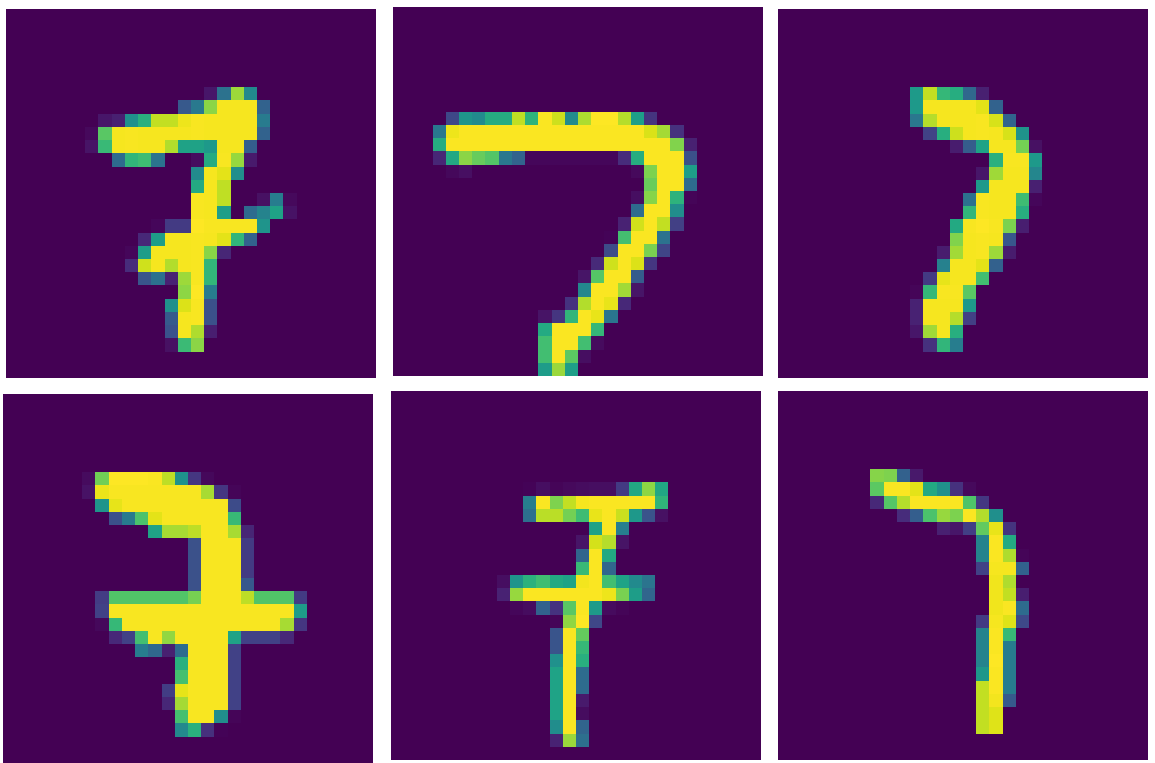}
\centering
\caption{The anomalous digit '7' recognized by OIAD in the unlabeled scenario for MNIST.}
\end{figure}
\begin{figure}[!htb]
\setlength{\abovecaptionskip}{-0.05cm}
\setlength{\belowcaptionskip}{-0.2cm}
\includegraphics[width=2.5in]{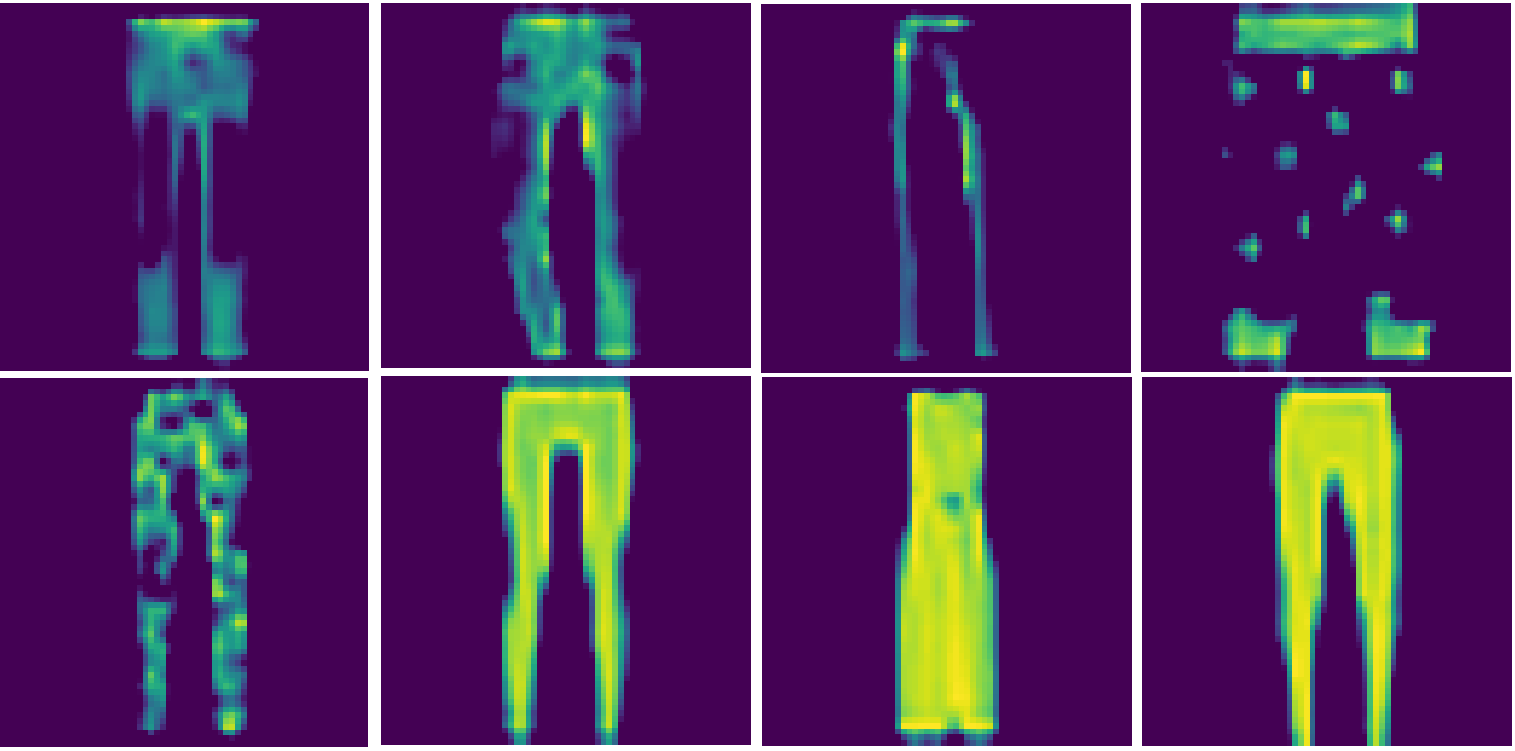}
\centering
\caption{The anomalous trousers recognized by OIAD in the unlabeled scenario for FMNIST.}
\end{figure}
\begin{figure}[!htb]
\setlength{\abovecaptionskip}{-0.05cm}
\setlength{\belowcaptionskip}{-0.2cm}
\includegraphics[width=3.5in]{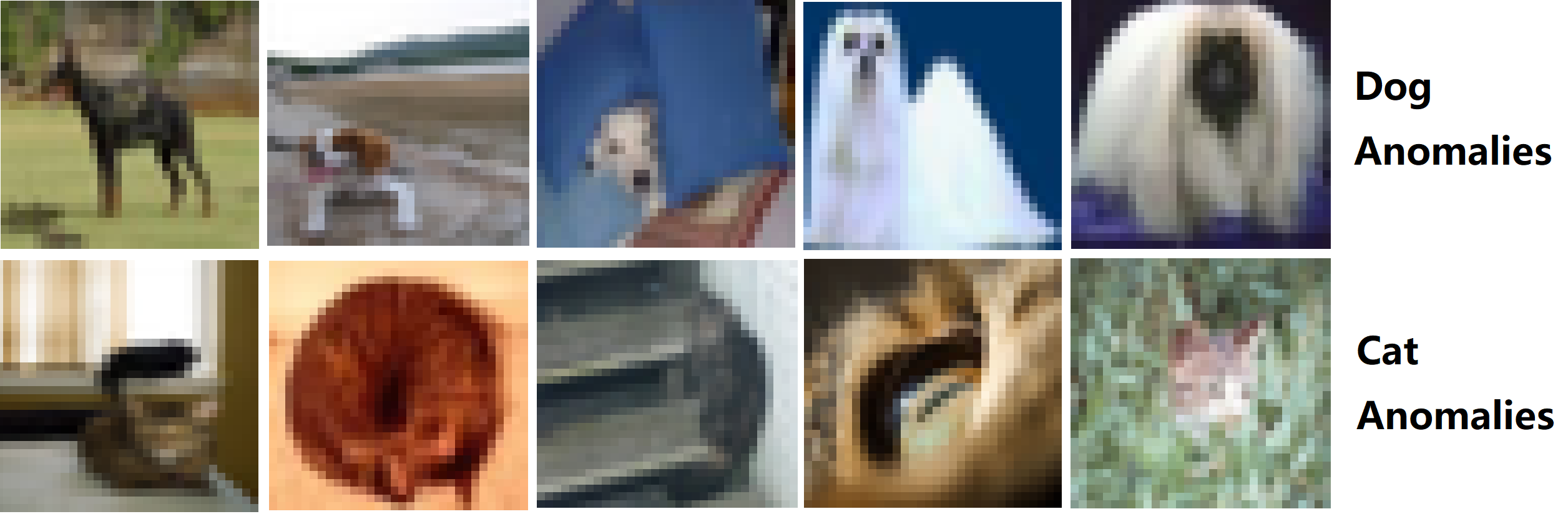}
\centering
\caption{The anomalous dogs/cats recognized by OIAD in the unlabeled scenario for CIFAR-10.}
\end{figure}
\section{Related Work}
Traditional non-parametric anomaly detection approaches include kernel density estimation (KDE) \cite{parzen1962estimation}, mixtures of Gaussians for active learning of anomalies \cite{pelleg2005active}, hidden Markov models for registering network attacks \cite{ourston2003applications}. However, non-parametric anomaly approaches suffer from the curse of dimensionality and are thus commonly insufficient for high-dimensional data. In addition, they are often fully-supervised, where a pre-processing step is needed to make the dataset balanced before applying any classification algorithm

For semi-supervised or unsupervised methods, most of the existing proposed approaches currently rely either on deep autoencoders or generative models.
Despite autoencoders have been primarily advanced for dimensionality reduction, they can be improved to anomaly detection problems. When an autoencoder is trained on normal instances, it will be trained to represent the main features in its latent space of normal instances. When an anomalous input is fed in the network, it is assumed it cannot be properly represented in the latent space, and thus the decoder reconstruction will be poor \cite{zhou2017anomaly, zhang2019deep}.
The other main approach is based on generative models, e.g. generative adversarial networks (GAN). When GAN is trained on normal data, the generator learns a “normality model” much like autoencoders do. If the generator is inverted, a comparison on the latent representations of normal and anomalous data can be used to detect anomalies \cite{akcay2018ganomaly,deecke2018image, schlegl2017unsupervised, zenati2018efficient}.

The main drawbacks of the existing deep learning anomaly detection are that (1) it is hard to estimate the data distribution in a tractable way; (2) well-labeled samples are required to train the anomaly detector; (3) the granularity of reconstruction error is hard to decide. Therefore, to the best of our knowledge, this is the first use of disentanglement learning and structure consistency for anomaly detection tasks. We believe that the ability of disentangled latent codes to create better fine granularity of reconstruction error evaluation can boost anomaly detection.

\section{Conclusion}
In this work, we proposed an OIAD technique based on disentanglement learning. The network is trained once on normal samples only without the requirement of anomalies, even the clean samples contain suspicious elements. Improving the disentanglement performance and feature-wise reconstruction evaluation are key ingredients to enhance the ability of anomaly detection of OIAD. We demonstrate that perturbing the disentangled latent space of the images can be leveraged for anomaly detection tasks. Experimental results show the OIAD has state-of-the-art performance, without any anomalies required and only based on threshold vectors that are easy to be decided experimentally.
As future work, we plan to investigate the use of OIAD for anomaly detection on text and time-series data.

\bibliographystyle{IEEEtran}
%

\bibliography{conference_101719}
\end{document}